%% file: lrec-coling2024-example.tex
\documentclass[10pt, a4paper]{article}

\usepackage[]{lrec-coling2024} 

\usepackage{times}
\usepackage{latexsym}

\usepackage[T1]{fontenc}

\usepackage[utf8]{inputenc}

\usepackage{times}

\usepackage{latexsym}
\usepackage{tabularx}
\usepackage{arydshln}

\usepackage[T1]{fontenc}

\usepackage[utf8]{inputenc}
\usepackage{CJK}
\newcommand{\jap}[1]{\begin{CJK}{UTF8}{min}#1\end{CJK}}
\usepackage{microtype}
\usepackage[normalem]{ulem}
\usepackage{comment}
\usepackage{booktabs}
\usepackage{threeparttable}
\usepackage{color,soul}
\usepackage{graphicx} 
\usepackage{tikz}
\usepackage{pgfplots}
\definecolor{DarkBlue}{RGB}{28,81,140}
\definecolor{DarkRed}{RGB}{151,11,28}
\definecolor{DarkGreen}{RGB}{49,147,70}
\definecolor{LightGreen}{RGB}{129,234,108}
\makeatletter
  \newcommand\greenuline{\bgroup\markoverwith{\textcolor{green}{\rule[-0.7ex]{2pt}{0.4pt}}}\ULon}
  \UL@protected\def\redwave{\leavevmode \bgroup 
    \ifdim \ULdepth=\maxdimen \ULdepth 3.5\p@
    \else \advance\ULdepth2\p@ 
    \fi \markoverwith{\lower\ULdepth\hbox{\textcolor{red}{\sixly \char58}}}\ULon}
\makeatother

\usepackage{comment}
\usepackage{times}
\usepackage{multicol}
\usepackage{multirow}
\usepackage{graphicx}
\usepackage{booktabs}
\usepackage{microtype}
\usepackage{algorithm}
\usepackage{algorithmic}

\usepackage{amsbsy}
\usepackage{amsmath,soul}
\usepackage{soulpos}

\usepackage{amsmath}
\usepackage{multirow}
\usepackage{makecell}
\usepackage{blindtext}

\usepackage{comment}
\usepackage{soulpos}
\usepackage{xcolor}

\usepackage[shortcuts]{extdash}
\usepackage{lipsum}

\usepackage{enumitem}

\newcommand{\mich}[1]{{\textcolor{black}{#1}}}

\newcommand\corpusname{NAIST-SIC}
\newcommand\alignedcorpusname{{\corpusname}-Aligned}

\title{
{\alignedcorpusname}: an Aligned English-Japanese Simultaneous Interpretation Corpus }

\name{\large Jinming Zhao$^{1,*,+}\thanks{* Work done while Jinming was a research intern at
NAIST.}$, Yuka Ko$^{2,+}$\thanks{+These authors contributed equally to this work.}, Kosuke Doi$^2$, Ryo Fukuda$^2$\\ \textbf{\large Katsuhito Sudoh$^2$, Satoshi Nakamura$^2$}}

\address{$^1$Monash University, Australia \\ $^2$Nara Institute of Science and Technology, Japan \\
jinming.zhao@monash.edu, \{ko.yuka.kp2, sudoh, s-nakamura\}@is.naist.jp \\
}

\abstract{
It remains a question that how simultaneous interpretation (SI) data affects simultaneous machine translation (SiMT). Research has been limited due to the lack of a large-scale training corpus. In this work, we aim to fill in the gap by introducing \emph{\alignedcorpusname}, which is an automatically-aligned parallel English-Japanese SI dataset. Starting with a non-aligned corpus {\corpusname}, we propose a two-stage alignment approach to make the corpus parallel and thus suitable for model training. The first stage is coarse alignment where we perform a many-to-many mapping between source and target sentences, and the second stage is fine-grained alignment where we perform intra- and inter-sentence filtering to improve the quality of aligned pairs. To ensure the quality of the corpus, each step has been validated either quantitatively or qualitatively. This is the first open-sourced large-scale parallel SI dataset in the literature. We also manually curated a small test set for evaluation purposes. Our results show that models trained with SI data lead to significant improvement in translation quality and latency over baselines. We hope our work advances research on SI corpora construction and SiMT. 
Our data can be found at \url{https://github.com/mingzi151/AHC-SI}.
 \\ \newline \Keywords{simultaneous interpretation, interpreting corpus, corpus construction} }

\begin{document}

\maketitleabstract

\input{sections/1-introduction}
\input{sections/2-method}

\input{sections/3-experiment}
\input{sections/5-conclusion}
\nocite{*}
\section{Bibliographical References}\label{sec:reference}

\bibliographystyle{lrec-coling2024-natbib}
\bibliography{lrec-coling2024-example}



\end{document}

%% file: sections/1-introduction.tex
\section{Introduction}
\label{sec:introduction}

Simultaneous interpretation (SI) is a task where an utterance is translated in real-time. Simultaneous machine translation (SiMT) systems should produce reasonably good translations with low latency~\cite{ma2019stacl,DBLP:conf/eacl/ArthurCH21}. Due to the lack of a large-scale SI corpus, most SiMT systems are trained with offline machine translation (MT) corpora which are often abundant with easy access~\cite{DBLP:conf/acl/ZhengZMH19}. Yet, MT data is different from SI data because of the difference between offline and online translation in nature. 
Despite some efforts~\cite{zhao2021not}, it has always remained a question on how and to what extent SI data affects an SiMT system.

\citet{zhao2021not} recently made a call for constructing SI corpora to learn interpreters' behaviours in modelling, but no corpus has been proposed as of now. This is mainly contributed by the fact that collecting and building parallel SI corpora is exceptionally costly and time-consuming~\cite{shimizu2013constructing}. Most existing corpora are in a small scale, mainly for testing purposes~\cite{zhao2021not,bernardini2016epic};  see Table~\ref{tab:corpora-list}\footnote{The pSp corpus introduced in~\citet{Paulik-and-Waibel-2009} replies heavily on time information. Its quality is unclear due to the lack of open-source accessibility.} for detailed statistics. In addition, aligning source and target sentences presents great challenges.
\citet{doi2021large} presented a large-scale document-level SI corpus, {\corpusname}, and conducted analyses on a small manually-aligned subset. 
However, sentences are not aligned within each document pair for most of the corpus, so it cannot be used for model training due to the lack of alignment.

\begin{table}
\centering
\small
\scalebox{0.85}{
\begin{tabular}{lcc}
\hline \textbf{Corpora} & \textbf{Language} & \textbf{Size} \\
\hline
\citet{Tohyama-et-al-2004} & En$\leftrightarrow$Jp & 182 hours \\
\citet{shimizu2014collection} & En$\leftrightarrow$Jp & 22 hours \\
\citet{doi2021large} & En$\leftrightarrow$Jp & 304.5 hours \\
\hline
\citet{pan2019chinese} & Zh$\leftrightarrow$En & 6M tokens \\
\citet{zhang2021bstc}~(dev/test) & Zh$\rightarrow$En & 3 hours  \\
\hline
\citet{Paulik-and-Waibel-2009} & En$\leftrightarrow$Es & 217 hours \\
\citet{bernardini2016epic} & En, Fr, It & 95K tokens \\
\citet{kunz2021heicic} & En$\leftrightarrow$De & 83 hours \\
\citet{zhao2021not} & En$\leftrightarrow$De & 1K sent. \\
\citet{machacek21_interspeech} & En$\rightarrow$De, Cs & 10 hours \\
\citet{wang2021voxpopuli} & 15 lang. & 17.3K hours \\
\citet{przybyl2022epic} & En, De, Es & 10K sent. \\
\hline
\end{tabular}}
\caption{Existing SI corpora.}
\label{tab:corpora-list}
\end{table}

This work aims to fill the gap by building a large-scale parallel English-Japanese SI corpus named \emph{\alignedcorpusname}.
We start with the non-parallel {\corpusname} and propose a two-stage pipeline alignment approach consisting of coarse and fine-grained alignment to make it parallel.
The initial stage is coarse alignment which involves identifying minimal groups of source and target sentences that are considered translations of each other.
The next stage is fine-grained alignment, where intra- and inter-sentence filtering techniques are applied over coarse-aligned pairs to improve data quality.
We validate each step either manually or automatically, to ensure the quality of the data.
Meanwhile, we compile a small-scale, manually curated SI test set for testing purposes.
We additionally summarize alignment challenges and findings to guide future SI corpus construction for other language pairs. Lastly, we build SiMT systems based on our corpus and show significant improvement over baselines in both translation quality and latency.

%% file: sections/2-method.tex
\section{Raw Corpus Construction}
\paragraph{Raw SI data collection}

We used a portion of SI data from {\corpusname} in this study, referred to as \textsf{SI$^{RAW}$}. This data comprises professional simultaneous interpreters' real-time interpretations of TED talks, which span a variety of topics from technology to entertainment.\footnote{\url{https://www.ted.com/}}
Interpreters with varying experience levels (S-rank: 15 years, A-rank: 4 years, B-rank: 1 year) participated in the interpretations.
Note that \textsf{SI$^{RAW}$} underwent manual transcription. We specifically used interpretations by S-rank and A-rank interpreters, denoted as \textsf{SI$^{Srank\_RAW}$} and \textsf{SI$^{Arank\_RAW}$}, respectively.
Overall, the quality of \textsf{SI$^{Srank\_RAW}$} surpasses that of \textsf{SI$^{Arank\_RAW}$}.

\paragraph{Manual subset alignment}

\citet{doi2021large} conducted sentence-level alignment on 14 talks (a subset of talks interpreted by all three rank interpreters) for analysis. This process involved manual alignment of source sentences with target sentences, referred to as \textsf{true-align}. We refer readers to the original paper for more details.

\paragraph{Error analysis}
After a manual investigation on \textsf{true-align}, we categorise issues with the SI data into two groups: \textit{under-translation} and \textit{mis-translation}.
\textit{Under-translation} occurs when interpreters unintentionally omit content due to memory overload.
Interpreters may also omit information intentionally, using tactics such as summarization, which is permissible. For instance, in the following interpretation, omitting "And if you drill into that" does not impede comprehension of the source sentence.
\mich{
\begin{itemize} [leftmargin=1.5mm,parsep=0pt,partopsep=0pt]
\item[]\textsf{Source}: And if you drill into that, it's especially the case for men.
\item[]\textsf{Interpretation}: \jap{特に、男の人はそうなんです。} {\small[This is especially true for men.]}
\end{itemize}}
\textit{Mis-translation} is the result of interpreters mistakenly rendering source sentences, often due to cognitive overload. The  interpretation below clearly misrepresents the source sentence. 
\mich{
\begin{itemize} [leftmargin=1.5mm,parsep=0pt,partopsep=0pt]
\item[]\textsf{Source}: I think we should make this even more explicit.
\item[]\textsf{Interpretation}: \jap{分かったと思います。} {\small[I think I understand.]}
\end{itemize}}

\section{Parallel Corpus Construction}
In this section, we present how we perform alignment on \textsf{SI$^{RAW}$}~(\S\ref{align}).
Then, we detail how we split data to train/dev/test~(\S\ref{split}), followed by describing manual compilation of  the test set~(\S\ref{test}).


\subsection{Alignment}\label{align}

We propose a two-stage alignment method that the first stage involves coarse alignment, grouping minimal source and target sentences to establish translations, and the second stage refines these pairs through fine-grained alignment. Both stages are validated quantitatively and qualitatively, with insights and findings shared at each step.

\subsubsection{Coarse Alignment}

\paragraph{Alignment}
Each talk consists of $M$ source sequences, $\left(e_{1}, e_{2}, \ldots, e_{M}\right)$, and $N$ target sequences, $\left(f_{1}, f_{2}, \ldots, f_{N}\right)$, where each $e$ is at the sentence level, each $f$ is at the chunk level, and $N > M$.
Note that there is no clear punctuation to group target sequences to sentence level, due to the nature of SI.
The first step is to detect groups between these sequences that are translations of each other.\footnote{We tested various sentence and chunk combinations but found the current setup, 
to be the most effective. This aligns with our intuition, as source-side punctuation aids in semantically grouping target chunks.
}  

Due to \textit{under-translation} and \textit{mis-translation}, some sequences lack corresponding translations, requiring deletion operations.
We used the vecAlign~\cite{thompson2019vecalign} sentence aligner, which suits the above purposes. VecAlign creates source and target language sentences, compares sentence embeddings computed by LASER~\cite{artetxe2019massively} using dynamic programming, and yields candidate pairs $<E, F>$ with the lowest cost.\footnote{Even when substituting LASER with other advanced encoders, it consistently outperforms them.}
Pairs with a cost higher than 1 or with empty $E$ or $F$ indicating no corresponding translation were removed.

\paragraph{Validation}
The outputs of this stage are validated both qualitatively and quantitatively.
Manual inspection reveals a relatively high alignment quality, particularly for \textsf{SI$^{Srank\_RAW}$}.
To perform quantitative evaluation, we use \textsf{true-align} to automatically generate coarse alignment from talks, denoted as \textsf{auto-align}, and then measure the extent to which \textsf{true-align} can be recovered.
For a given source sentence, $e$, we measure the similarity between the automatically aligned target sentence, $F^a$, and the manually aligned sentence, $F^m$, as follows:
\begin{equation}
\setlength{\abovedisplayskip}{3pt}
\setlength{\belowdisplayskip}{3pt}
\setlength{\abovedisplayshortskip}{3pt}
\setlength{\belowdisplayshortskip}{3pt}
    s =\frac{{LCS}(F^a, F^m)}{|F^m|}
\end{equation}
where $LCS (A, B)$ stands for the longest common substring between strings $A$ and $B$.

We define $f^a$ as correctly aligned to $e$ when the similarity measure $s$ exceeds a threshold $\epsilon$. Higher $\epsilon$ values result in lower accuracy.
Notably, with $\epsilon$ set to 0.8, the overall accuracy for the S-rank subset in \textsf{true-align} remains at 80\%, underscoring vecAlign's effectiveness in addressing the issue.
We name the output at this stage \textsc{Coarse}, and we also prepare S-rank data as \textsc{Coarse$^{Srank}$}.


\paragraph{Finding} 
After the manual check of the outputs, we summarise our main findings as follows.
\setlist[itemize]{align=parleft,left=0pt..1em}
\begin{itemize}
\item The quality of alignment results largely depends on the quality of interpretation data.
\textsf{SI$^{Srank\_RAW}$} has a better quality of data than \textsf{SI$^{Arank\_RAW}$}, and thus the aligned pairs of the former have higher quality. 
\item The quality of interpretation data is influenced by talk difficulty, which can be assessed from two perspectives: the speaker and the interpreter. A talk becomes challenging when the speaker speaks rapidly, employs jargon, has an accent, etc. It is also contingent on the interpreter's skills and background knowledge, etc.
\item Interpreters may exhibit varying performance levels for different talks. For easy talks, their performance remains consistent, contrasting the typical phenomenon of performance decline due to cognitive load over time. We attribute this to the interpreters working in a less stressful, simulated environment. In the case of difficult talks, the expected performance decline is observed.
\end{itemize}

\paragraph{Human Evaluation}
We further analysed half of a talk from \textsf{SI$^{Srank\_RAW}$}, denoted as \textsc{sample} (comprising of 78 pairs).
Our findings indicate that 82\% of the pairs are well-aligned, 11\% are reasonably aligned, and the remainder are poorly aligned. 
Within the 11\% subset, misalignment often occurs at the beginning/end of the target sentence, and removing it would enhance alignment quality.\footnote{Occasionally, the misaligned part should belong to the previous or the next sentence. We leave the potential improvement to future work.} 
While these statistics may vary across talks, 
these observations guide our alignment design.

\subsubsection{Fine-grained Alignment}\label{fine}
The second stage is to perform fine-grained alignment consisting of two filtering stages.
The purpose is to improve the accuracy of aligned pairs.

\paragraph{Intra-sentence filtering} 
Based on our observations, we decide to filter out the beginning and last chunk of a target sentence if it does not carry any substantial meaning.
For example, \jap{じゃあ} (meaning ``then'' in English) often appears as a filer in Japanese and should be removed.
We consider Japanese words conveying important information (i.e., content words): ``NOUN'', ``PROPN'', ``PRON'', ``VERB'' or ``NUM''.
Manual inspection of the output from this step on \textsc{sample} shows that this simple heuristic accurately detects and removes chunks that contain no content words in most of the time.
We call the resulting data together the S-rank subset \textsc{Intra} and \textsc{Intra$^{Srank}$}.  
\paragraph{Inter-sentence filtering} 
To address potential quality disparities in \textsc{Coarse} alignment for talks other than \textsc{Sample}, we introduce rigorous surface-level and semantic-level filtering rules. 
i) For an aligned pair <$E$, $F$>, we calculate $\alpha$, representing the percentage of content words in $E$ covered by $F$.
Content words include "NOUN," "PROPN," and "NUM,"; "VERB" is excluded due to its perceived lesser importance during interpretation~\cite{seeber2001intonation}. This accounts for interpreters' techniques aimed at managing cognitive load, allowing for certain word omissions.
ii) We compute the length ratio $\gamma$ between $F$ and its corresponding offline translation $T$, generated via Google Translate from $E$.
Large or small $\gamma$ indicate potential \textit{mis-translation} and \textit{under-translation} issues.
iii) We measure semantic similarity, $\eta$, between $F$ and $T$ with BLEURT~\cite{sellam2020bleurt}, a semantic-based MT metric.
As $T$ is the correct translation of $E$, high $\eta$ implies high coverage of $E$ by $F$.
We denote the entire data and S-rank data as \textsc{Inter} and \textsc{Inter$^{Srank}$}, respectively.  
\begin{table}[t]
\centering
    \scalebox{0.85}{
    \begin{tabular}{ l|c|c| c  }
    \hline
    \thead{Data} & Subset & \# Talks & \# Pairs \\
    \hline
    \textsc{Coarse} & train & 831  & 67,079   \\
    \textsc{Coarse$^{Srank}$}& train & 472&  41,597 \\
    \textsc{Intra}& train &831 &  66,834  \\
    \textsc{Intra$^{Srank}$}& train  & 472& 41,436 \\
    \textsc{Inter}& train & 831& 50,096  \\
    \textsc{Inter$^{Srank}$} & train & 472 &32,039 \\
    \textsc{AUTO-DEV} & dev & 10 & 732 \\
    \textsc{AUTO-TEST} & test & 15 & 1,176 \\
    \textsc{SI$^{DEV}$} & dev & 4 & 238 \\
    \textsc{SI$^{TEST}$} & test & 5 & 383 \\
    \hline
    \end{tabular}}
    \caption{Dataset statistics}
    \label{tab:data}
\end{table}

\paragraph{Validation}
Each of the above steps has been manually validated by one of the authors.
That said, the optimal values for $\alpha$, $\gamma$ and $\eta$ vary by talk, and we leave automatically learning optimal values per talk for future work. 

\subsection{Data Split}\label{split}
Both our data and MuST-C~\cite{di2019must}, a common dataset in SiMT (both speech and text), contain TED Talks.
To prevent potential data contamination, we ensured our train set (831 talks) is present in Must-C train data. We created training set variations for experimentation.
For evaluation, we selected 10 and 15 talks from \textsc{Intra} as our dev and test sets, referred to as \textsc{AUTO-DEV} and \textsc{AUTO-TEST}. Importantly, these sets do not intersect with MuST-C training data. Additionally, we curated manual dev and test sets for testing.

\mich{\subsection{Dev \& Test Set Curation}}\label{test}
To guarantee the quality of the dev and test sets, we randomly selected four and five talks interpreted by S-rank interpreters for annotation for the subsets.\footnote{\mich{The annotators are two PhD students with backgrounds in interpreting and language analysis. Agreement on the annotations was reached between them in cases of ambiguity.}}  Data curation involves a two-step process.
\paragraph{Label aligning}
We observed that a correctly aligned pair may not be suitable for evaluation due to issues such as \textit{under-translation} or \textit{mis-translation}. For instance, in the following pair, the target SI sentence aligns correctly with the source sentence but is not a correct translation.

\begin{itemize} [leftmargin=1.5mm,parsep=0pt,partopsep=0pt]
\item[]\textsf{Source}: They haven't given up on government.
\item[]\textsf{Interpretation}: \jap{政府は、諦めていないのです。} {\small[The government hasn't given up.]}
\end{itemize}

\mich{Another instance involves words with similar pronunciations, especially for proper nouns. While hearing them in a speech may be appropriate, using them in written text could lead to errors. Hence, we instructed the annotators to assign two labels: \textit{good\_align} for well-aligned pairs, and \textit{good\_mt} for suitable for evaluation.  Assigning the \textit{good\_align} label is generally straightforward and less ambiguous. However, assigning the \textit{good\_mt} label requires a judgement call; annotators collaborated on this task to reach a consensus and minimize bias. It is unrealistic to expect perfect interpretations; therefore, the principle for annotation dictates that omitting important content (e.g., numbers and proper nouns) is not allowed, but omitting less important content is permissible.} 

\mich{For the above example, the labels would be True and False, respectively.} 

\paragraph{Sentence editing}
We further asked annotators to perform two sequential tasks on the target sentence. (1) Rectify alignment issues from automatic alignment. \mich{This is mainly done by introducing "true COARSE" for the test set if chunks are removed from or re-introduced to the target sentence as a result of auto alignment. We note that the need of merging sentences is rare; even in the case of merging, the resulting sentences do not lead to good pairs. This can be explained by human's behavior that when interpreters have to merge sentences, they make mistakes more easily.} (2) Perform minimal manual post-editing to ensure faithfulness and reliability without altering the data distribution significantly. \mich{Examples include, but not limited to, replacing pronouns with referred entities, removing self-correction to improve fluency, and correcting numbers.} 

In the example below, the annotator would add the text highlighted in red to the target sentence.
\begin{itemize} [leftmargin=1.5mm,parsep=0pt,partopsep=0pt]
\item[]\textsf{Mod\_Interpretation}: \jap{\textcolor{red}{彼らは}政府\textcolor{red}{を}諦めていないのです。} {\small[They haven't given up on government.]} 
\end{itemize}
The entire process, consuming approximately 20 hours, resulted in \textsc{SI$^{TEST}$}. 
Additionally, we manually annotated three talks for the dev set, creating \textsc{SI$^{DEV}$}. Data statistics are provided in Table \ref{tab:data}.

%% file: sections/3-experiment.tex
\section{Experiments}

\subsection{Experimental settings}

\paragraph{Datasets} In addition to our proposed data shown in Table \ref{tab:data}, we also complemented it with the Must-C v2 En-Ja dataset,
which is an offline dataset with 328,639 training instances. 

\paragraph{baselines} We applied test-time wait-k~\cite{ma2019stacl} on offline MT models trained on offline (i.e., Must-C) and various online datasets, including \textsc{Coarse}, \textsc{Coarse$^{Srank}$}, \textsc{Intra} and \textsc{Intra$^{Srank}$}, as our baseline models. 

\begin{figure}[t]
  \centering
    \includegraphics[scale=0.35] {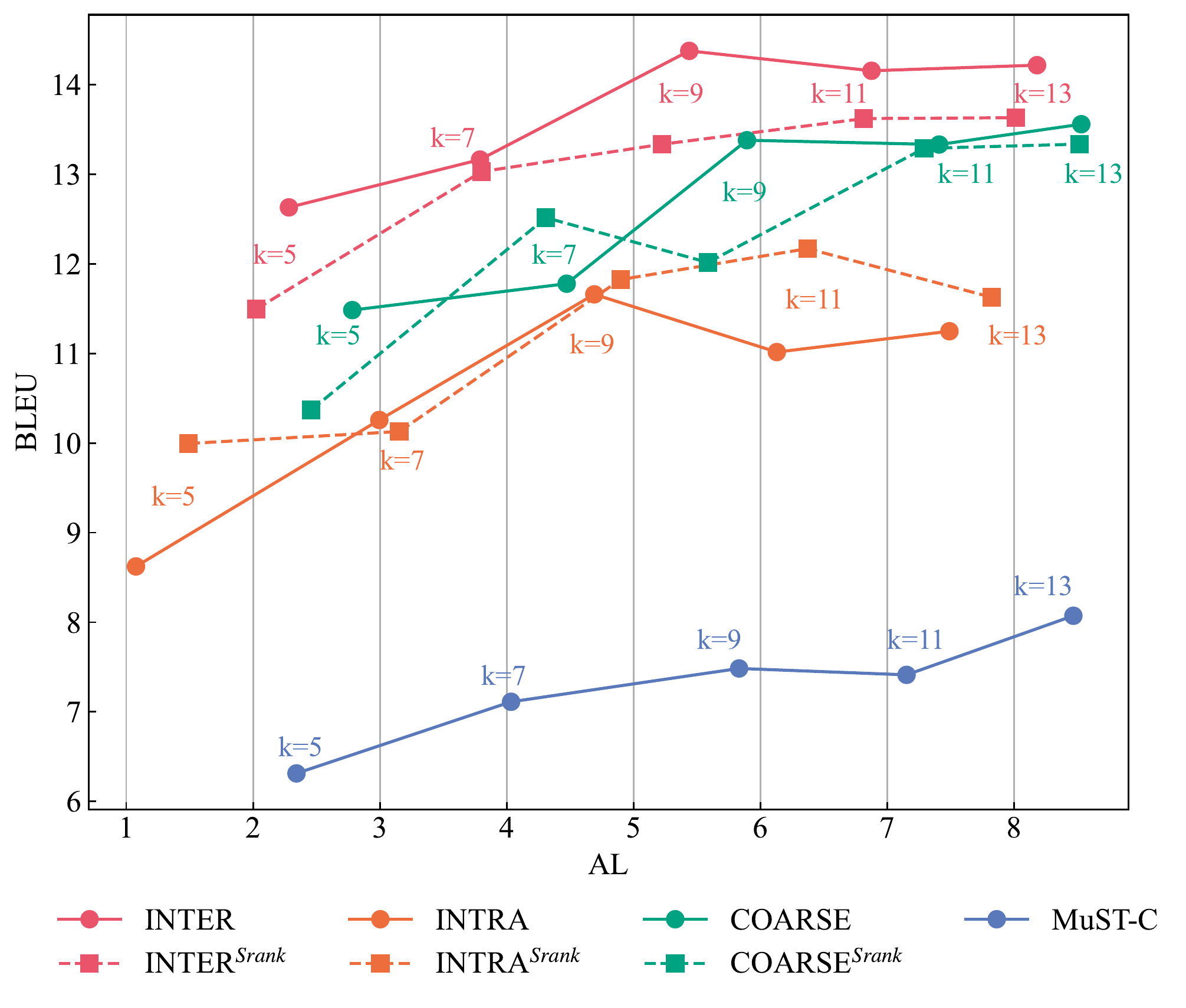}
  \caption{Translation quality and latency for wait-k systems trained on Must-C and various SI data. 
  }
  \vspace{-0.2cm}
  \label{fig:main}
\end{figure}

\vspace{-0.2cm}
\paragraph{Implementation details} For the wait-k systems, we primarily followed the fairseq toolkit~\cite{ott2019fairseq} instructions with some distinctions. 
We used separate vocabularies for English and Japanese with SentencePiece~\cite{kudo-richardson-2018-sentencepiece} which was trained on Must-C training data. 
We set the vocabulary size to 8,000, as larger or smaller sizes yielded worse results. We used a batch size of 7,168 with the update frequency of 4, and set the dropout rate on 0.3. Early-stopping occurred after 16 epochs if validation loss stagnated. We selected the best checkpoint based on the loss on \textsc{SI$^{DEV}$} for all SI experiments. We evaluated model performance on \textsc{SI$^{TEST}$} with the SimulEval toolkit~\cite{ma2020simuleval} on \textsc{SI$^{TEST}$} where translation quality is measured in BLEU\footnote{\url{https://github.com/mjpost/sacrebleu}} and latency in average lagging (AL)~\cite{ma2019stacl}. We also measured translation quality with BLEURT.
All models start from an offline MT system trained with MuST-C. 

\vspace{-0.2cm}
\subsection{Results}
Figure \ref{fig:main} shows BLEU and AL scores in a set of k values (i.e., 5, 7, 9, 11, 13).
SiMT systems trained \textsc{Inter} and \textsc{Inter$^{Srank}$} outperform models trained on Must-C significantly, by an average of 6.43 and 5.74 BLEU scores, respectively, across all latency settings. This demonstrates the effectiveness of SI data despite Must-C En-Ja having significantly more data than our corpus.  
BLEURT scores confirm that both \textsc{Inter} and \textsc{Inter$^{Srank}$} offer the highest translation quality.\footnote{The performance order of other systems differs with BLEU and BLEURT. We attribute this to imperfect measurement metrics for SiMT, which is beyond the scope of this work.} 

%% file: sections/5-conclusion.tex
\section{Conclusion}
\label{sec:conclusion}
The question of how simultaneous interpretation (SI) data impacts simultaneous machine translation (SiMT) remains unresolved. This is due to the lack of large-scale SI training data, constructing which imposes great challenges. In this work, we filled in the gap by introducing {\alignedcorpusname}, an automatically aligned English-Japanese SI corpus, together with a small-scale human annotated SI test set. We proposed a two-stage alignment approach to align source data with target SI data. Our results show that systems trained on our proposed data surpassed baselines by a large margin. We share findings and insights with SI corpus, hoping to offer guidance on future research on SI corpus construction. 

\section*{Acknowledgements}
Part of this work was supported by JSPS KAKENHI Grant Number JP21H05054.